\documentclass{article}
\usepackage{spconf,amsmath,graphicx,subfigure}
\usepackage{hyperref}
\usepackage{amssymb}
\usepackage{array,enumitem}

\makeatletter
\def\blfootnote{\xdef\@thefnmark{}\@footnotetext}
\makeatother

\title{Cloud Detection from RGB Color Remote Sensing Images With Deep Pyramid Networks}
\name{Savas Ozkan, Mehmet Efendioglu, Caner Demirpolat}
\address{TUBITAK Space Technologies Research Institute 
\\ Remote Sensing Group
\\ Ankara, Turkey
\\ \{savas.ozkan, mehmet.efendioglu, caner.demirpolat\}@tubitak.gov.tr}

\begin{document}
%
\maketitle
\begin{abstract}

Cloud detection from remotely observed data is a critical pre-processing step for various remote sensing applications. In particular, this problem becomes even harder for RGB color images, since there is no distinct spectral pattern for clouds, which is directly separable from the Earth surface. In this paper, we adapt a deep pyramid network (DPN) to tackle this problem. For this purpose, the network is enhanced with a pre-trained parameter model at the encoder layer. Moreover, the method is able to obtain accurate pixel-level segmentation and classification results from a set of noisy labeled RGB color images. In order to demonstrate the superiority of the method, we collect and label data with the corresponding cloud/non-cloudy masks acquired from low-orbit Gokturk-2 and RASAT satellites. The experimental results validates that the proposed method outperforms several baselines even for hard cases (e.g. snowy mountains) that are perceptually difficult to distinguish by human eyes.

\end{abstract}

\vspace{-0.3cm}
\section{Introduction}
\vspace{-0.3cm}
\label{sec:intro}

\blfootnote{Project Website: \url{https://github.com/savasozkan/cloud_detection}}
The presence of clouds due to climate factors limits the clear acquisition of content information from the Earth surface for almost all optical sensors. Ultimately, this reduces the visibility and affects adversely the processing of data for many remote sensing applications such as classification, segmentation and change detection etc. Hence, detection/elimination of cloudy coverages constitutes an important pre-processing step for remote sensing.

In particular, RGB color bands are more sensitive to these atmospheric scattering conditions compared to the high wavelength sensors (i.e. infrared/multi-spectral)~\cite{low2014}. Thus, this problem becomes even harder and the spatial content of the image needs to be leveraged rather than singly spectral properties of clouds as in multi-spectral/infrared sensors. For this reason, addressing the problem from the perspective of object segmentation and classification can yield more intuitive results. Moreover, more generalized solutions, i.e. instead of sensor-specific rules/thresholds, can be presented~\cite{salient2015, multi2017}.

In this paper, we tackle the cloud detection problem by presenting a framework based on deep pyramid network architecture (DPN)~\cite{u2015, feature2016}. Compared to the existing rule-based methods~\cite{object2012, char2006, auto2016, auto2015}, the proposed method exploits texture information exhibited from cloudy/non-cloudy pixels with high-level features. This improves classification decisions without the need of any specific spectral information, since a pre-trained encoder network is capable of extracting rich and distinct high-level representations for visual objects in the images. Moreover, due to the architecture, the network is concurrently optimized for both segmentation and classification phases. Lastly, since the ground truth cloud masks are quite noisy (i.e. achieving perfect pixel-level annotations is quite difficult~\cite{thesis2013}), use of a pre-trained model for the abstract representation of an input provides robustness to the overall segmentation and classification phases.

Rest of the paper is organized as follows: related works are reviewed in Section 2. Section 3 is reserved for the detail of the proposed method and the problem statement. Experimental results, dataset and baseline methods are explained in Section 4 and the paper is concluded in Section 5.

\vspace{-0.3cm}
\section{Related Work}
\vspace{-0.3cm}
\label{sec:format}

In this section, we review the literature for RGB color satellite images as well as other optical sensors such as multi-spectral/infrared to demonstrate the complexity of the problem for visible domain. 

The methods used for multi/infrared bands are frequently based on radiometric properties of clouds/surface as reflectance and temperature.~\cite{object2012, char2006} exploit the variations of reflectance in thermal bands to distinguish clouds from the surface. Harb et. al.~\cite{auto2016} propose a processing chain based on the thermal pattern of clouds with morphological filtrations.  Similarly, Braaten et. al.~\cite{auto2015} extend the assumption to multi-spectral data. However, these methods highly depend on sensor models (i.e. since they are rule-based methods) and the derived solution cannot be generalized to different sensors by using similar assumptions for band information. 

Differently, multi-temporal methods aim to detect clouds based on background changes in time by which data is acquired in different time-instances. Zhu et. al.~\cite{auto2014} combine the thermal cloud patterns with time-series data to detect more accurate cloud masks. Moreover, the method~\cite{mask2017} uses temporal data to estimate clouds with a non-linear regression algorithm. However, the main limitation of the methods is that the time series of data are assumed to possess smooth variations on ground surfaces while abrupt changes for clouds. Furthermore, recording such dense data practically increase the operational cost.

In order to generalize the solution, classification-based approaches learn a set of parameters from training samples to distinguish clouds from the surface. Hu et. al.~\cite{salient2015} extract several low-level features such as color, texture features etc. to estimate pixel-level masks. Recently,~\cite{multi2017} classifies locally sampled patches (i.e. by a Super-Pixel (SP) algorithm) with a Convolutional Neural Network (CNN) as cloud or non-cloud. 

\vspace{-0.3cm}
\section{Cloud Detection With Pyramid Networks}
\vspace{-0.3cm}
\label{sec:pagestyle}

As mentioned, since there is no explicit spectral/physical pattern for clouds in RGB color satellite images, we treat the problem as an object segmentation and classification problem in order to make a realistic problem formulation.

 In particular, the texture details around cloudy regions indicate distinct visual patterns for detection/segmentation phases. Our aim is to extract high-level abstract representations from data and iteratively merge them to make pixel-level classification decisions. Moreover, the proposed method is able to compute segmentation and classification phases concurrently to optimize the network in an end-to-end learning manner, thus there is no need to employ these layers separately as in~\cite{multi2017}.

\begin{figure}[t]
\centering
\vspace{-0.2cm}
\includegraphics[scale=0.22]{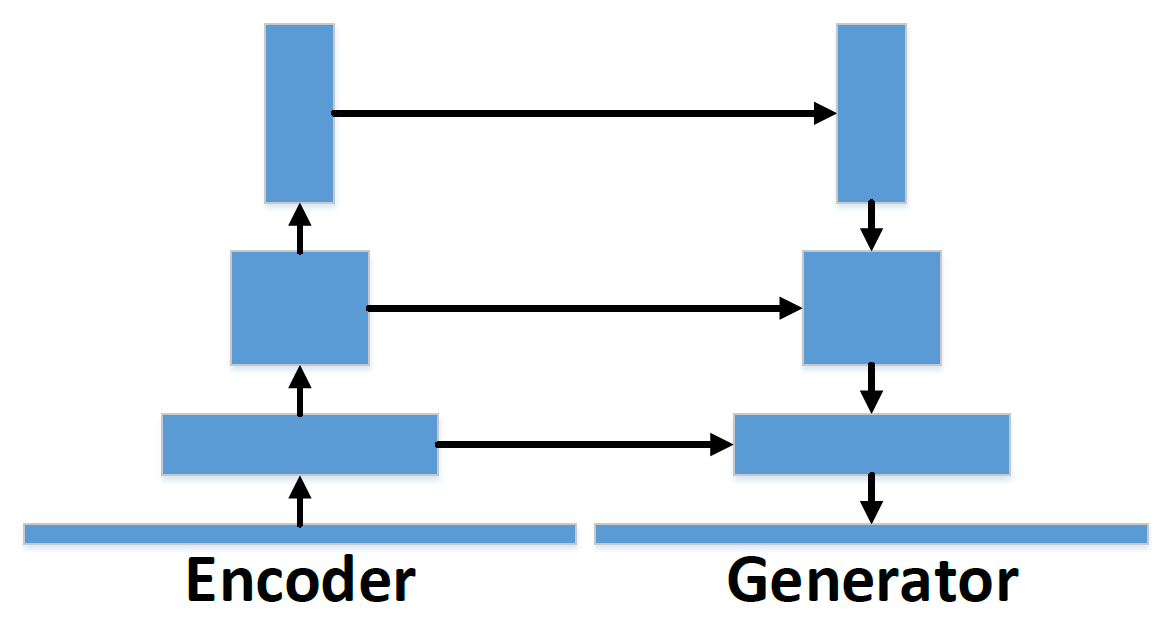}
\vspace{-0.5cm}
\caption{Deep Pyramid Network with Encoder and Generator filter blocks.}
\vspace{-0.5cm}
\label{fig:flow}
\end{figure}

\vspace{-0.4cm}
\subsection{Formulation}
\vspace{-0.2cm}
\label{ssec:subhead}

Suppose we are given a RGB color satellite image $\bold{x} \in \mathbb{R}^{W \times H \times 3}$ and the method aims to generate an image mask $\bold{y} \in \mathbb{R}^{W \times H \times 2} $ that implicitly corresponds to two channel classification decisions for ground surface and cloud/haze coverages.

Therefore, the main objective is to learn a set of parameters $\theta_c$ and $\theta_g$ for encoder and generator functions $C(.)$ and $G(.)$ such that the input-target error should be minimized for a set of training pairs $\{\bold{x}_i, \bold{y}_i \}$ based on a loss function:
\begin{eqnarray}
\label{eqn:lossc}
\mathcal{L} = -\dfrac{1}{N} \sum_{i} \bold{y_i} \log({ \bold{p_i} ) }
\end{eqnarray}

\noindent
where $\bold{p_i} = G( C(\bold{x}_i, \theta_c), \theta_g )$ is the mask prediction of the network for the input $\bold{x}_i$. The softmax cross-entropy loss in Eq. 1 maximizes the similarity of optimum input-target transformation. Moreover, $N$ corresponds to the mini-batch size. In the inference stage, the decisions of cloudy/non-cloudy coverages are computed based on the outputs of these learned functions.

\vspace{-0.2cm}
\subsection{Architecture}
\vspace{-0.2cm}

Our deep network architecture consists of two main filter blocks~\cite{u2015, feature2016}. First, encoder block $C(.)$ extracts robust abstract representations from a RGB color image. Then, generator block $G(.)$ computes pixel-level segmentation and classification masks according to the responses of the encoder block. The overall architecture is illustrated in Figure 1.

\noindent
\textbf{Encoder}: Encoder block takes an image as input and iteratively computes abstract representations by down-sampling responses. Practically, the goal of the block is to unveil distinct patterns about data which assist the generator so as to obtain an optimal image mask. Moreover, information flows to the generator are maximized with skip-connections~\cite{residual2016}.

Throughout the paper, we experimented with two different encoder models:

\begin{itemize}[leftmargin=0.4cm,rightmargin=0cm]
\vspace{-0.2cm}
\item First, a model with 5 convolutional layers and random parameter initialization is used. At each layer, we employ a batch normalization layer and an activation function, i.e. ReLU, after a convolution layer. Later, we down-sample the responses with stride 2. However, we found out that the random initialization lacks to reach an optimal solution due to the fact that ground truth unwillingly contains noisy labels by omission and/or registration noise during labeling~\cite{thesis2013}. This ultimately affects adversely the parameters at the end and the parameters (i.e. $\theta_c$) tend to generate false-alarms in the inference stage. 

\vspace{-0.2cm}
\item We use the convolutional responses of a pre-trained model, i.e. `conv1\_2', `conv2\_2', `conv3\_2', `conv4\_2' and `conv5\_2' in VGG-19~\cite{vgg2014}, and no finetuning is allowed for the encoder layers. Eventually, this mitigates the problem and more confident responses are obtained for an input.  Note that even if the parameters of the model are trained for a different object recognition problem, the studies have already shown that it is still capable of attaining best accuracies on several remote sensing applications~\cite{high2017, deepfeat2016}.

\end{itemize}

\noindent
\textbf{Generator}: At each layer, the generator block fuses the abstract representations extracted by the encoder block by adding and up-sampling (with factor $2\times$) recursively as illustrated in Fig. 2. Similarly, we use batch normalizations and ReLU functions at the layers to speed up the optimization. Other advantage of these functions is to improve the sparsity of the responses as explained in~\cite{end2017} for remote sensing.

At the last layer, we utilize a softmax activation to produce classification decisions, i.e. cloud or ground surface, thus it is inclined to set the decisions to either 0 or 1 for the masks at the end of the learning stage.

\vspace{-0.2cm}
\subsection{Implementation Details}
\vspace{-0.2cm}

As a pre-processing step, we first normalize each pixel in an image with the constant value computed in~\cite{vgg2014}, even if a pre-trained encoder model is used or not. Ultimately, it centers data to zero-mean space and data becomes reproducible for the pre-trained model.

For the parameter optimization, Adam optimizer~\cite{adam2014} with momentum $\beta_1 = 0.9$ and $\beta_2 = 0.999$ is used and the training rate is set to 0.0001. Moreover,  the value of $N$ is determined as 10 for $512 \times 512$ RGB color images and maximum mini-batch iteration is set to 20K.  Note that no data augmentation is utilized throughout the training stage. Lastly, all codes are implemented on Python using Tensorflow framework. The models are trained/evaluated on NVIDIA Tesla K40 GPU card.

\begin{figure}[t]
\centering
\vspace{-0.7cm}
\includegraphics[scale=0.18]{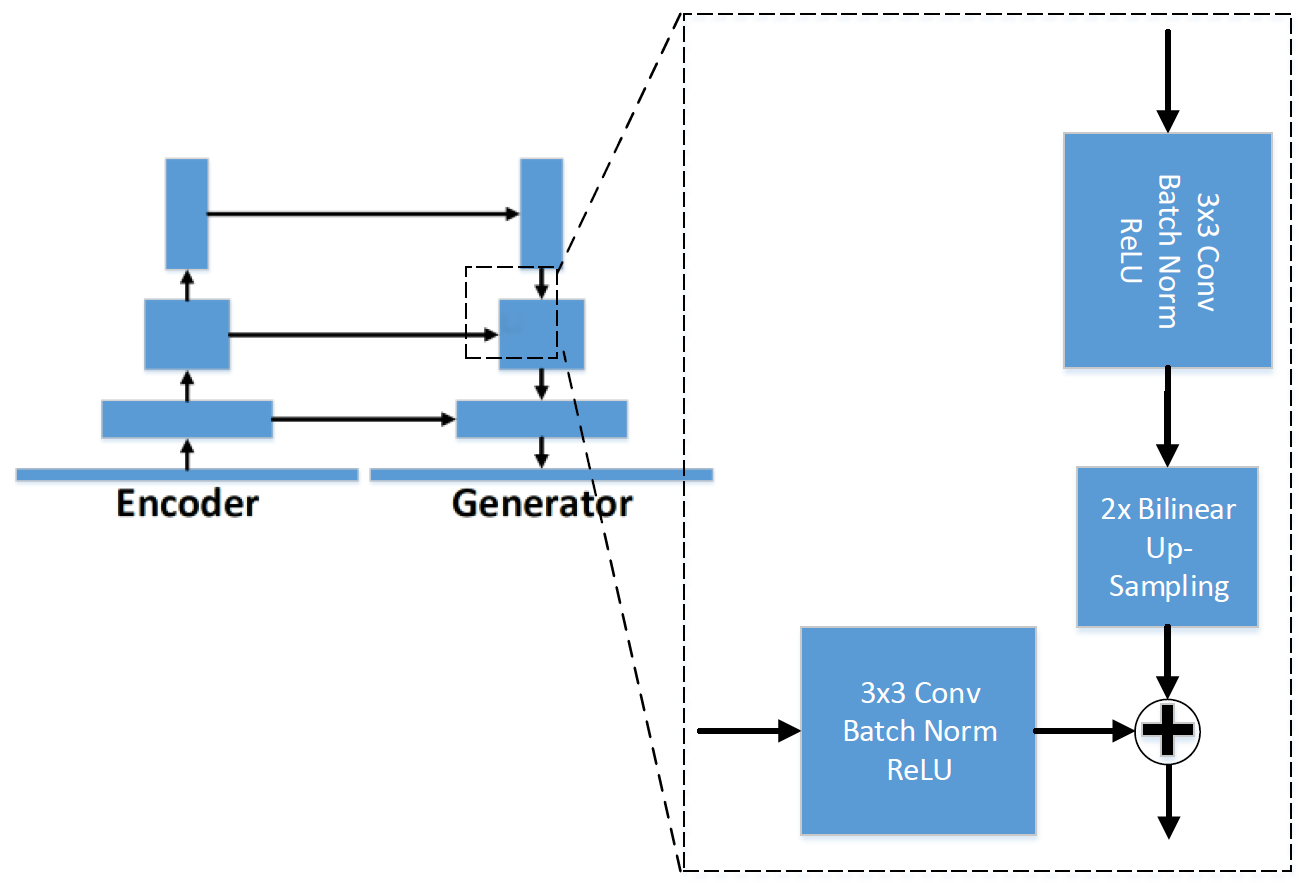}
\vspace{-0.5cm}
\caption{Detail visualization of Generator filter block at a single layer.}
\vspace{-0.5cm}
\label{fig:flow}
\end{figure}

\vspace{-0.3cm}
\section{Experiments}
\vspace{-0.3cm}
\label{sec:typestyle}

In this section, we mention the details of the dataset we used in the experiments. Later, we report/discuss the experimental results conducted on this dataset.

\vspace{-0.2cm}
\subsection{Dataset}
\vspace{-0.2cm}

The dataset consists of 20 images acquired from low-orbit RASAT and Gokturk-2\footnote{\url{https://gezgin.gov.tr/}} satellites, and their RGB resolutions are 15.0 m and 5.0 m respectively~\cite{gok2016}. In particular, we opt to use the outputs of two different sensors in the dataset to demonstrate the generalization capacity of the proposed method. Moreover, Level-1 processed data is utilized in the experiments to reduce the defects caused by platform motion and optical distortion. The ground truth masks are manually labeled by human experts. Lastly, all methods are trained on 15 images and the rest is reserved for the testing stage.

\vspace{-0.2cm}
\subsection{Experimental Results}
\vspace{-0.2cm}

To evaluate the success of the proposed method, we compare the method with two baselines, deep pyramid network (DPN) and the combination of CNN with Super-Pixel segmentation as in~\cite{multi2017}. Moreover, performance is measured by three score metrics, namely \textit{Accuracy} (correctness of the prediction), \textit{Precision} (reliability of the prediction) and \textit{Latency} (inference time).

We report the performance scores in Table 1. From the results, the proposed method (i.e. DPN+VGG-19) achieves best accuracy and precision scores. Particularly, our method significantly improves the precision score. This stems by the fact that replacing a pre-trained parameter model at the encoder block provides robustness to noisy-labeled data in the learning phase and it ultimately reduces the false-alarm in the inference stage. Another reason is that the proposed method is able to achieve segmentation and classification phases concurrently, thus the parameters are optimized by this way to estimate best segmentation masks rather than employing these steps separately as in~\cite{multi2017}. Lastly, this also provides some advantages in the computation time (i.e on CPU for 3583$\times$3584 resolution) as reported in Table 1 (Note that ~\cite{multi2017} needs to generate a decision with CNN for each local patch). 

Furthermore, we illustrate the classification masks of the proposed method and SP+CNN~\cite{multi2017} for the test images in Fig. 3\footnote{Note that you can find the results of all methods as well as ground truth masks in the project webpage with better visual quality.}. Perceptually, our method obtains impressive results particularly for hard cases such as snowy mountains. Moreover, the method is also able to detect haze coverages (i.e. the last column in Fig. 3), even though there is a limited number of training samples for such haze type in the dataset. The reason is that the network exploits the texture around clouds rather than color information, since their patterns are more discriminative for clouds compared to snow/saturated cases.

\begin{figure*}[t]
\centering
\vspace{-0.8cm}
\subfigure{
\label{fig:a}
\includegraphics[scale=0.14]{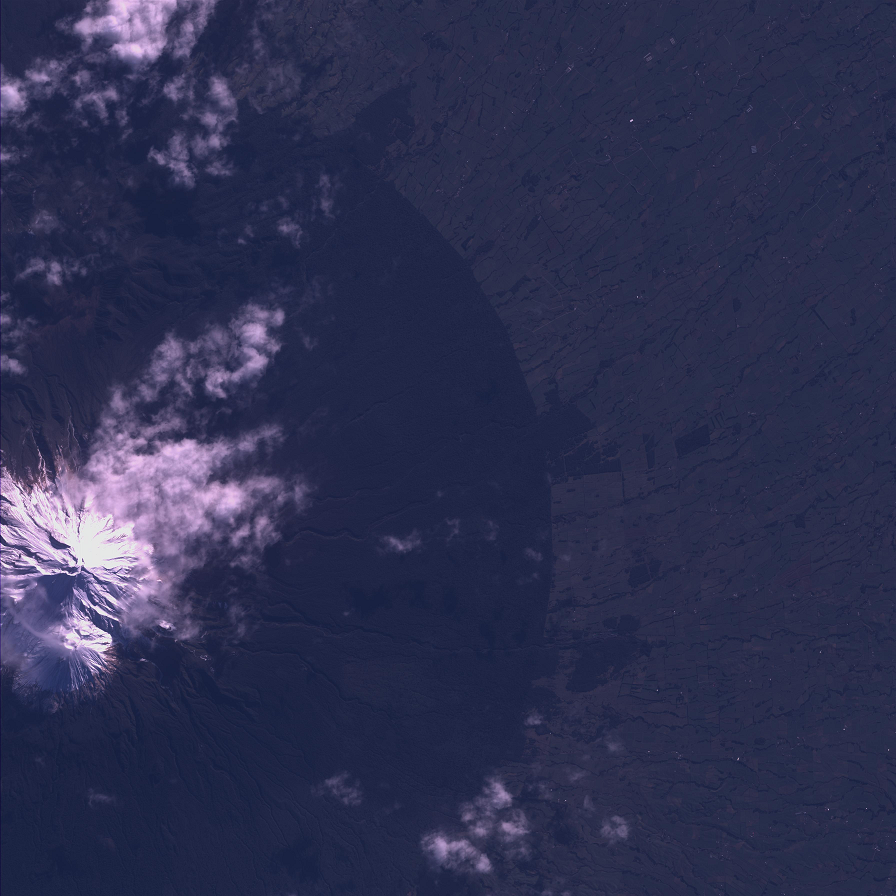}}
\subfigure{
\label{fig:b}
\includegraphics[scale=0.14]{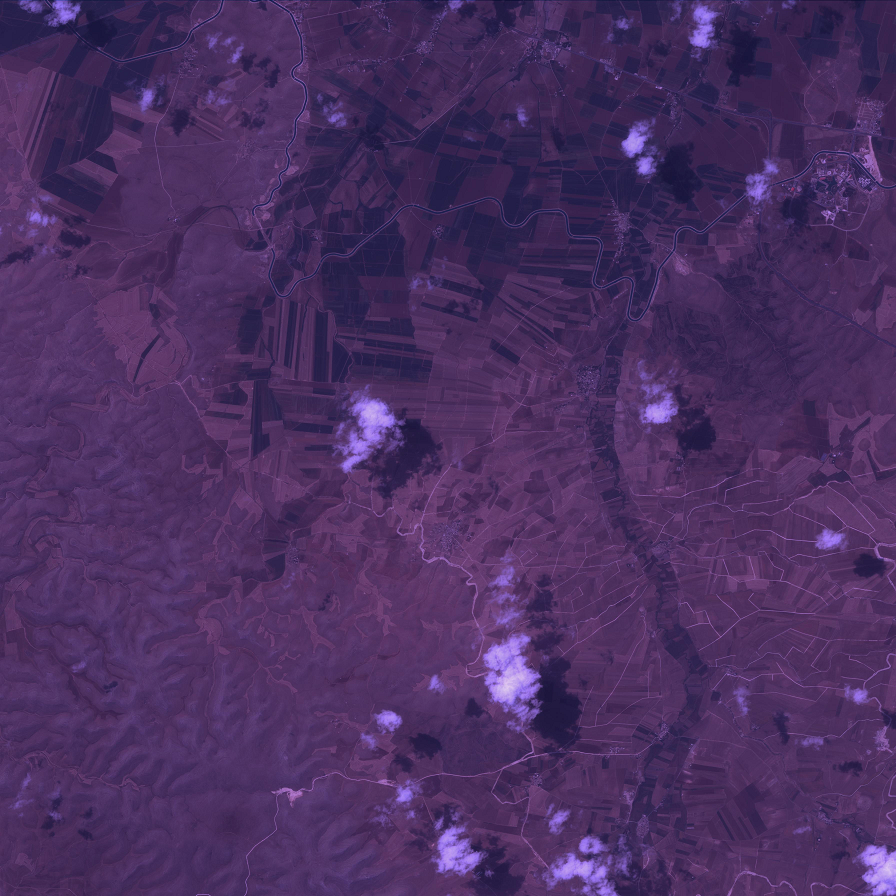}}
\subfigure{
\label{fig:c}
\includegraphics[scale=0.14]{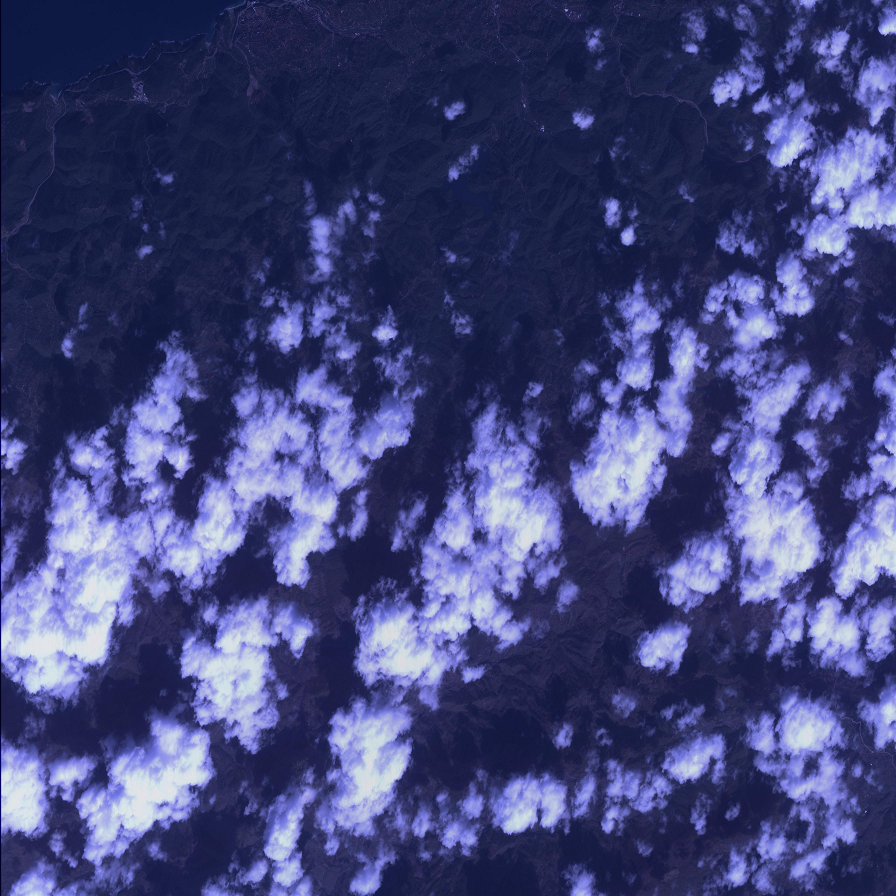}}
\subfigure{
\label{fig:d}
\includegraphics[scale=0.14]{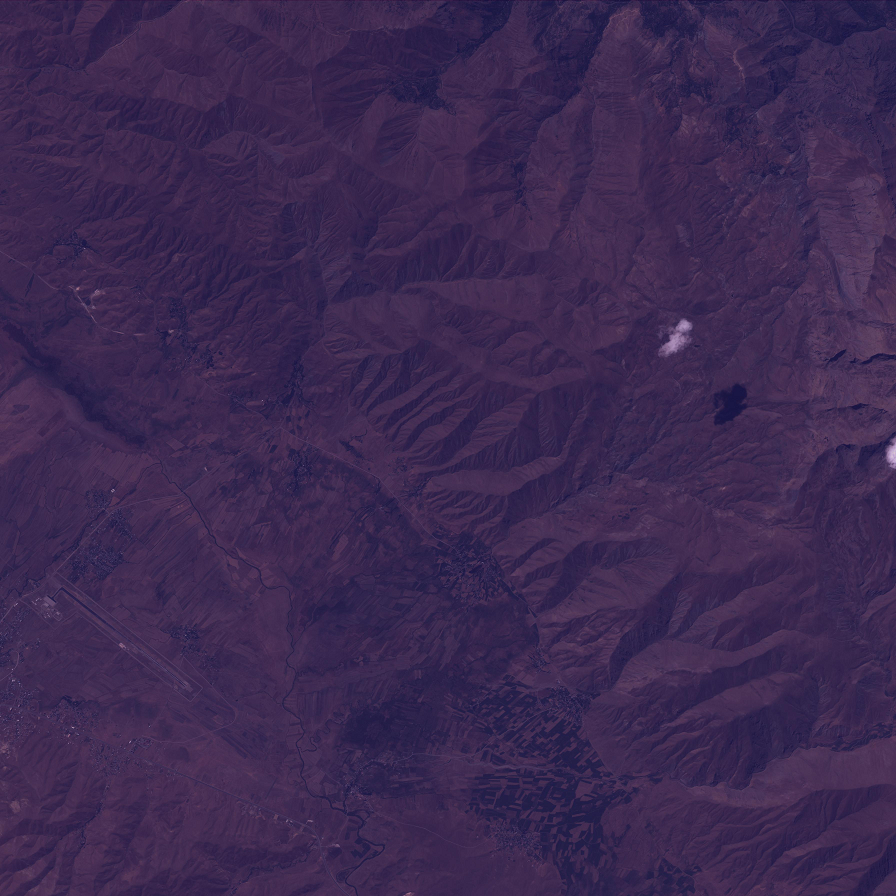}}
\subfigure{
\label{fig:d}
\includegraphics[scale=0.14]{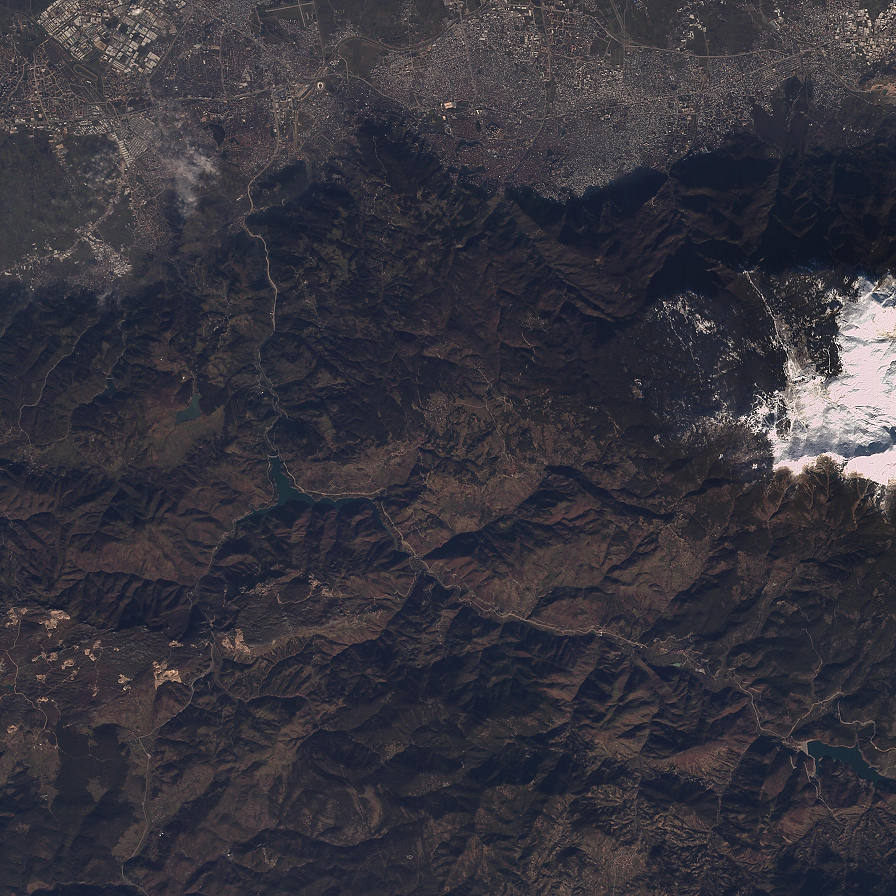}}

\vspace{-0.2cm}
\subfigure{
\label{fig:a}
\includegraphics[scale=0.14]{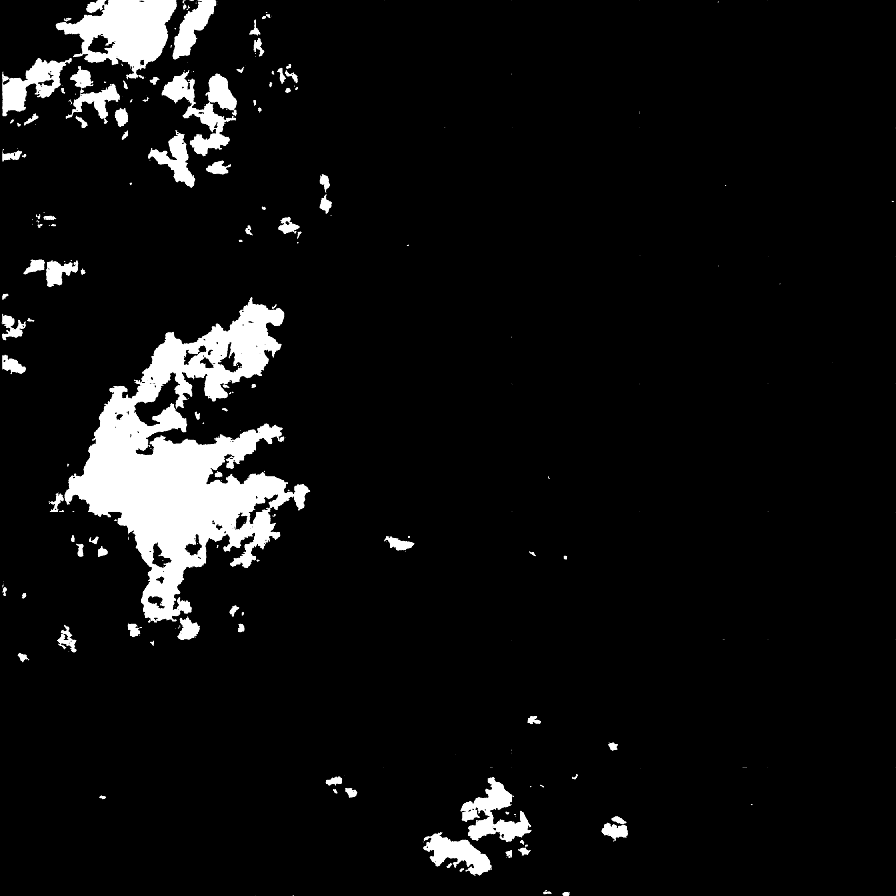}}
\subfigure{
\label{fig:b}
\includegraphics[scale=0.14]{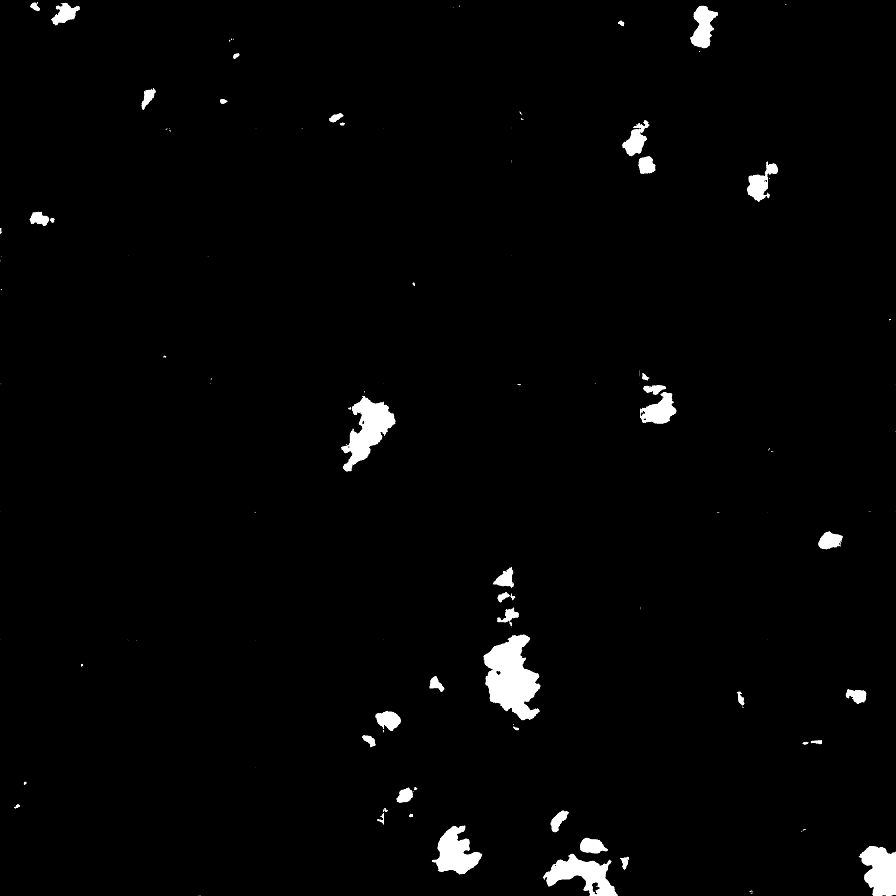}}
\subfigure{
\label{fig:c}
\includegraphics[scale=0.14]{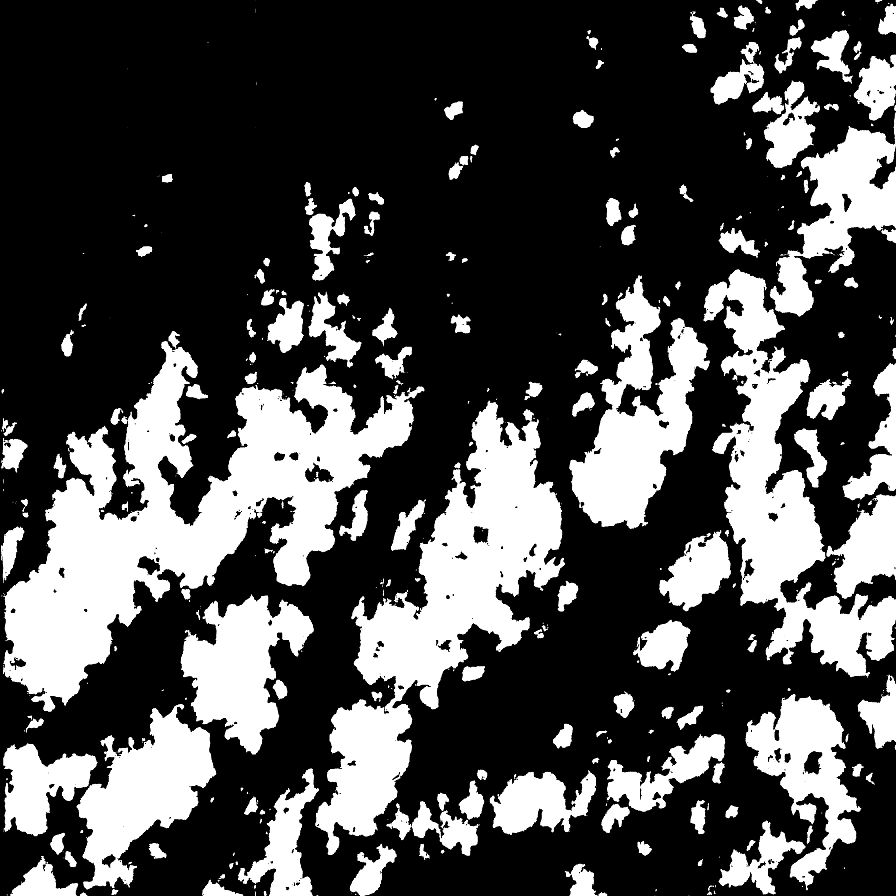}}
\subfigure{
\label{fig:d}
\includegraphics[scale=0.14]{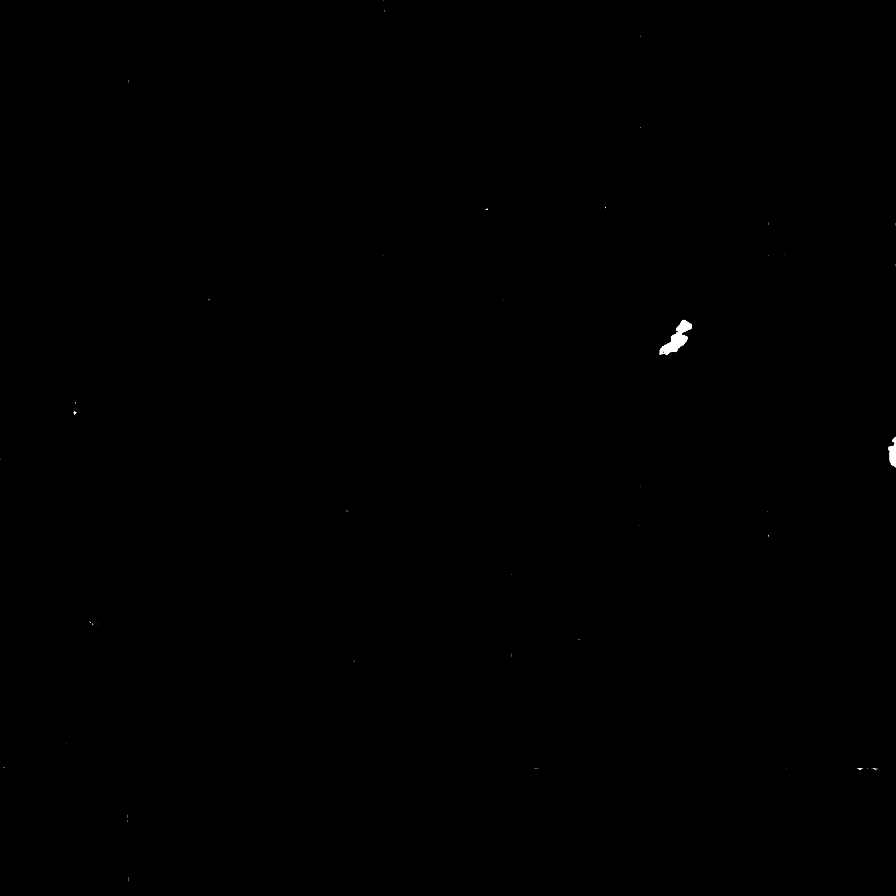}}
\subfigure{
\label{fig:d}
\includegraphics[scale=0.14]{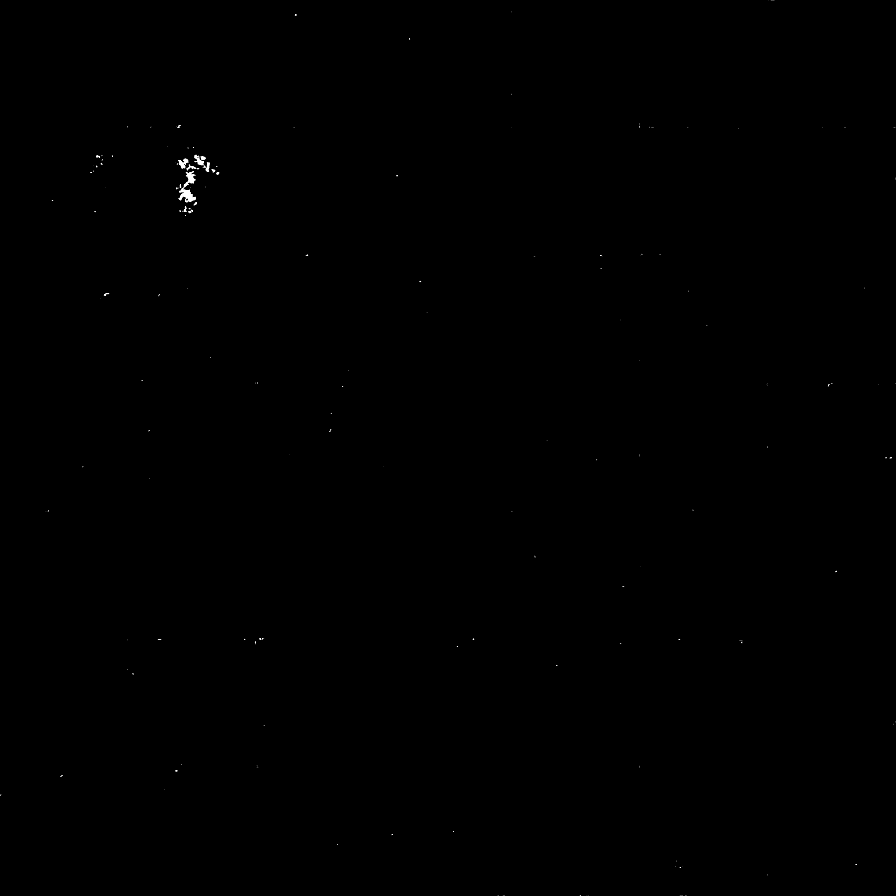}}

\vspace{-0.2cm}
\subfigure{
\label{fig:a}
\includegraphics[scale=0.14]{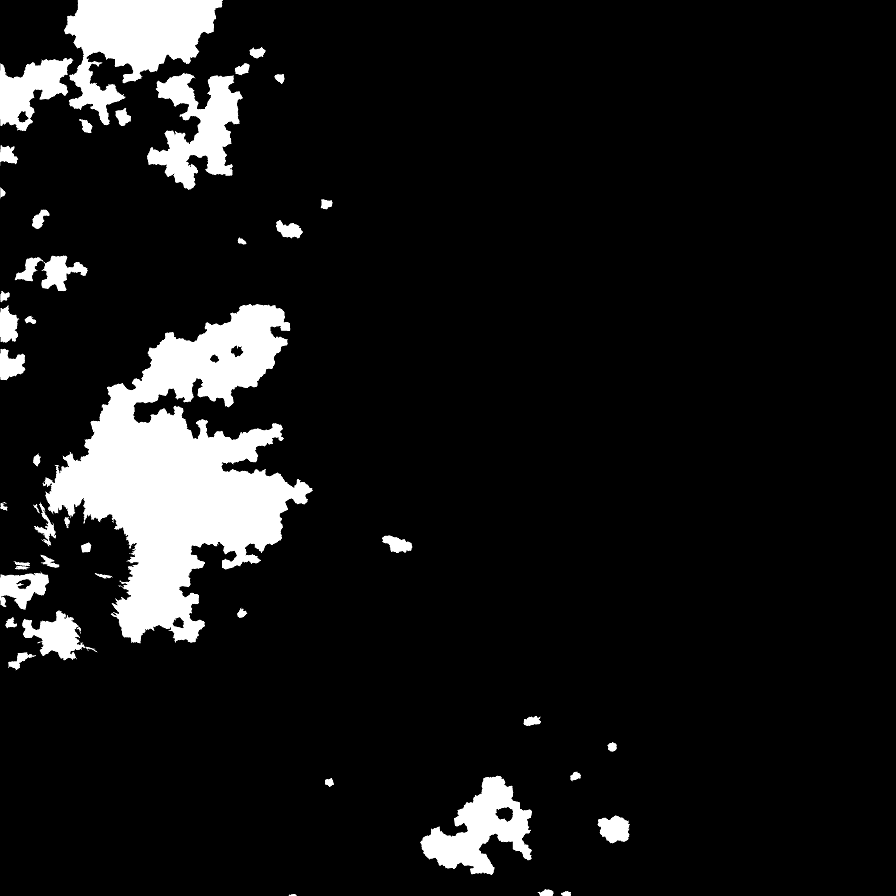}}
\subfigure{
\label{fig:b}
\includegraphics[scale=0.14]{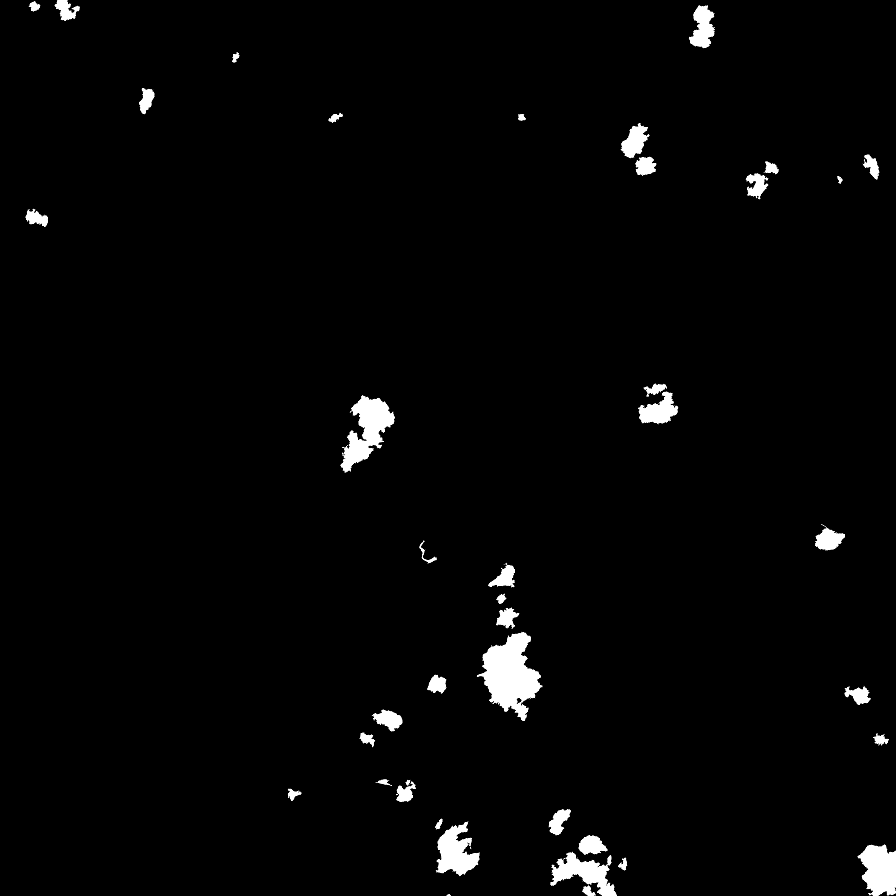}}
\subfigure{
\label{fig:c}
\includegraphics[scale=0.14]{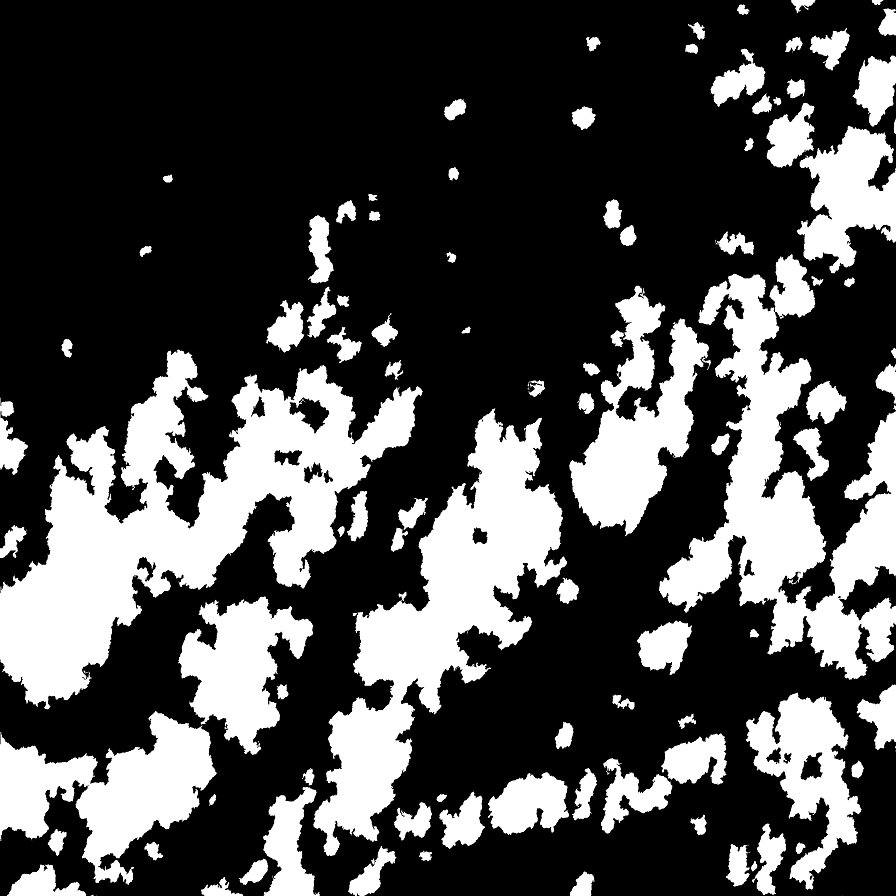}}
\subfigure{
\label{fig:d}
\includegraphics[scale=0.14]{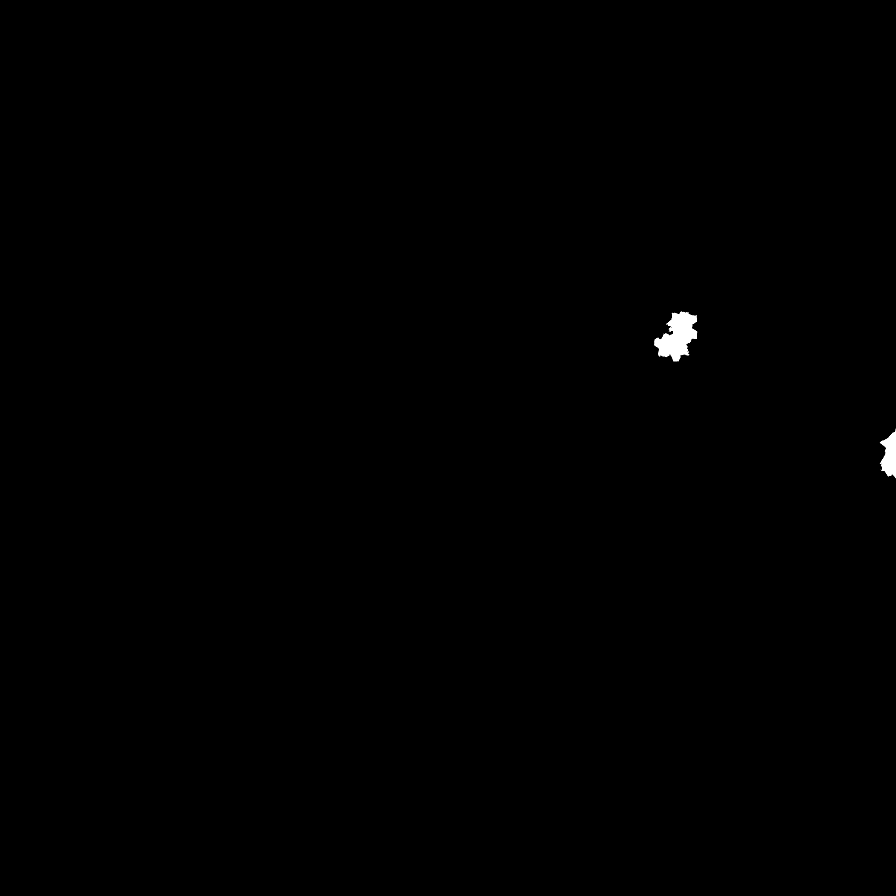}}
\subfigure{
\label{fig:d}
\includegraphics[scale=0.14]{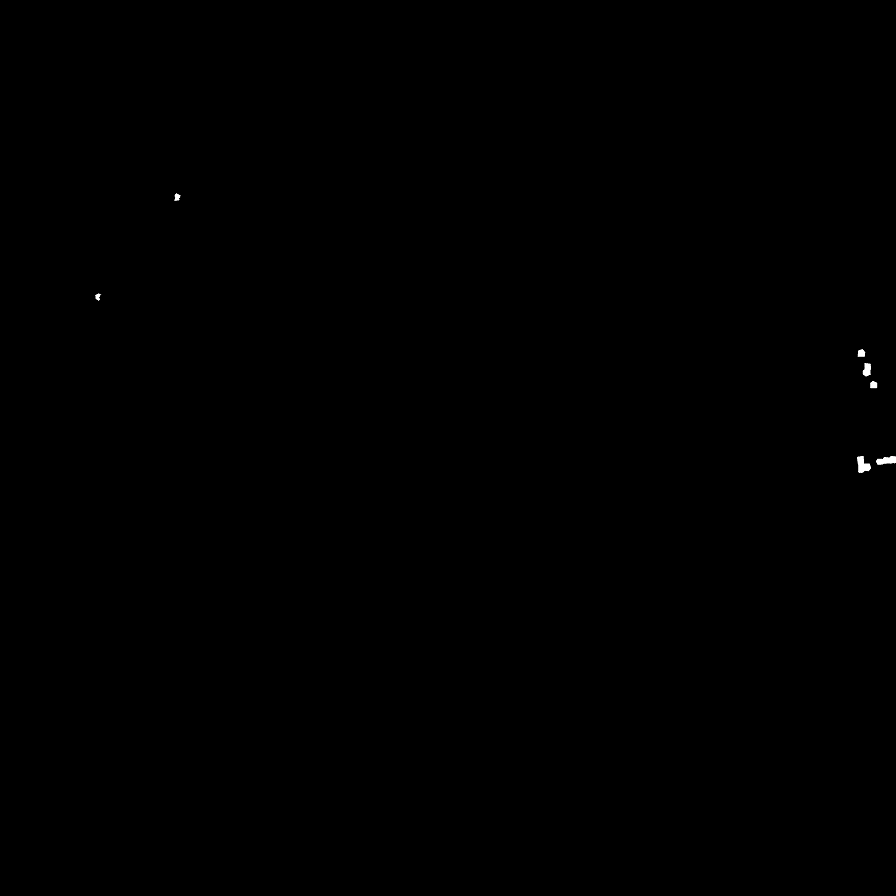}}

\vspace{-0.3cm}
\caption{Cloud masks for DPN+VGG-19 (second row) and SP+CNN~\cite{multi2017}(third  row) with RGB color images (first row).}
\vspace{-0.5cm}
\label{fig:data1}
\end{figure*}

\begin{table}[t]
\begin{center}
\vspace{-0.5cm}
\begin{tabular}{ l c c c }
\hline \hline
Image Id & Accuracy & Precision & Latency \\ 
\hline 
SP+CNN~\cite{multi2017}         & 0.9820 & 0.6676 & $\approx$30 min. \\ 
DPN                & 0.9815  & 0.7502 & $\approx$1 min. \\ 
\hline
DPN+VGG-19 (ours)     & \textbf{0.9874}  & \textbf{0.8776} & $\approx$1 min. \\ 
\hline \hline
\end{tabular}

\vspace{-0.2cm}
\caption{Accuracy and Precision scores for three methods. In particular, DPN+VGG-19 significantly improves the precision score compared to other methods. Latency of the methods is also reported for the inference stage.}
\vspace{-0.5cm}
\label{tab:result}
\end{center}
\end{table}

\vspace{-0.4cm}
\section{Conclusion}
\vspace{-0.3cm}
\label{sec:majhead}

In this paper, we propose a deep pyramid network to tackle cloud detection from RGB color images. The method is able to generate pixel-level decisions by exploiting spatial texture information about visual data. Moreover, we show that the integration of a pre-trained CNN model at the encoder layer improves the accuracy of classification masks, since more confident hidden representations are extracted from noisy labeled data. From the experimental results, the proposed methods quantitatively outperforms the baselines and obtains perceptually superior results on the dataset.

\vspace{-0.4cm}
\section{Acknowledgments}
\vspace{-0.3cm}
\label{sec:print}

The authors are grateful to NVIDIA Corporation for the donation of Tesla K40 GPU card used for this research. 

\bibliographystyle{IEEEtran}
\bibliography{Template}

\end{document}